\documentclass{article}
\usepackage{uai2020}
\usepackage[margin=1in]{geometry}

\usepackage{times}
\usepackage{microtype}
\usepackage{mathrsfs}
\usepackage{graphicx}
\usepackage{subfigure}
\usepackage{amssymb}
\usepackage{booktabs} 
\usepackage{bm}
\usepackage{url}            
\usepackage{booktabs}       
\usepackage{amsfonts}       
\usepackage{nicefrac}       
\usepackage{microtype}      
\usepackage{bm}
\usepackage{amsmath}
\usepackage{amsfonts}
\usepackage{amssymb}
\usepackage{mathtools}
\usepackage{titlesec}
\usepackage{wrapfig}
\usepackage{caption}
\usepackage{cite}
\usepackage{enumitem}
\usepackage[utf8]{inputenc} 
\usepackage[T1]{fontenc}
\usepackage{setspace}

\usepackage{fancyhdr}
\usepackage{color}
\usepackage{algorithm}
\usepackage{algorithmic}
\usepackage{natbib}
\usepackage{eso-pic} 
\usepackage{forloop}
\definecolor{mydarkblue}{rgb}{0,0.08,0.45}
\usepackage[colorlinks,linkcolor=blue,anchorcolor=black,citecolor=mydarkblue]{hyperref}

\usepackage{array}
\newcommand{\PreserveBackslash}[1]{\let\temp=\\#1\let\\=\temp}
\newcolumntype{C}[1]{>{\PreserveBackslash\centering}p{#1}}
\newcolumntype{R}[1]{>{\PreserveBackslash\raggedleft}p{#1}}
\newcolumntype{L}[1]{>{\PreserveBackslash\raggedright}p{#1}}

\usepackage{hyperref}

\title{\LARGE Accelerating Reinforcement Learning with a Directional-Gaussian-Smoothing Evolution Strategy
}

 \author{ {\bf \large Jiaxin Zhang}\thanks{ \; Computer Science and Mathematics Division,
 Oak Ridge National Laboratory,
 Oak Ridge, TN 37831.}
\qquad
 {\bf \large  Hoang Tran}\thanks{\; Computer Science and Mathematics Division,
 Oak Ridge National Laboratory,
 Oak Ridge, TN 37831.} 
\qquad
 {\bf \large  Guannan Zhang}\thanks{\; Corresponding author, Computer Science and Mathematics Division,
 Oak Ridge National Laboratory,
 Oak Ridge, TN 37831. Email: zhangg@ornl.gov.}
 }

\begin{document}

\maketitle

\begin{abstract}
Evolution strategy (ES) has been shown great promise in many challenging reinforcement learning (RL) tasks, rivaling other state-of-the-art deep RL methods. Yet, there are two limitations in the current ES practice that may hinder its otherwise further capabilities. First, most current methods rely on Monte Carlo type gradient estimators to suggest search direction, where the policy parameter is, in general, randomly sampled. Due to the low accuracy of such estimators, the RL training may suffer from slow convergence and require more iterations to reach optimal solution. Secondly, the landscape of reward functions can be deceptive and contains many local maxima, causing ES algorithms to prematurely converge and be unable to explore other parts of the parameter space with potentially greater rewards. In this work, we employ a Directional Gaussian Smoothing Evolutionary Strategy (DGS-ES) to accelerate RL training, which is well-suited to address these two challenges with its ability to i) provide gradient estimates with high accuracy, and ii) find nonlocal search direction which lays stress on large-scale variation of the reward function and disregards local fluctuation. Through several benchmark RL tasks demonstrated herein, we show that DGS-ES is highly scalable, possesses superior wall-clock time, and achieves competitive reward scores to other popular policy gradient and ES approaches.    

\end{abstract}

\section{INTRODUCTION}
\label{sec:intro}
Reinforcement learning is a class of problems which aim to find, through trial and error, a feedback policy that prescribes how an agent should act in an uncertain, complex environment to maximize some notion of cumulative reward \citep{Sutton-RLbook98}. Traditionally, RL algorithms have mainly been employed for small input and action spaces, and suffered difficulties when scaling to high-dimensional problems. With the recent emergence of deep learning, powerful non-linear function approximators such as deep neural networks (DNN) can be integrated into RL and extend the capability of RL in a variety of challenging tasks which would otherwise be infeasible, ranging from playing Atari from pixels \citep{Mnih15,Mnih16}, playing expert-level Go \citep{Silver-Nature16} to robotic control \citep{andrychowicz2017hindsight,Lillicrap-DDPG}.

Among the most popular current deep RL algorithms are Q-learning methods, policy gradient methods, and evolution strategies. Deep Q-learning algorithms \citep{Mnih15} use a DNN to approximate the optimal Q function, yielding policies that, for a given state, choose the action that maximizes the Q-value. Policy gradient methods \citep{Sehnke10}
improve the policies with a gradient estimator obtained from sample trajectories in action space, examples of which are A3C \citep{Mnih16}, TRPO \citep{Schulman_TRPO} and PPO \citep{Schulman-PPO}. 

This work concerns RL techniques based on evolution strategies. ES refers to a family of blackbox optimization algorithms inspired by ideas of natural evolution, often used to optimize functions when gradient information is inaccessible. This is exactly the prominent challenge in a typical RL problem, where the environment and policy are usually nonsmooth or can only be accessed via noisy sampling. It is not totally surprising ES has become a convincing competitor to Q-learning and policy gradient in deep RL in recent years. Unlike policy gradient, ES perturbs and performs policy search directly in the parameter space to find an effective policy, which is now generally considered to be superior to action perturbation \citep{SigaudStulp13}. The policy search can be guided by a surrogate gradient \citep{salimans2017evolution}, completely population-based and gradient-free \citep{Such-GA17}, and hybridized with other exploration strategies such as novelty search and quality diversity \citep{Conti_NIPS18}. It has been shown in those works that ES is easy to parallelize, and requires low communication overhead in a distributed setting. More importantly, being able to achieve competitive performance on many RL tasks, these methods are advantageous over other RL approaches in their high scalability and substantially lower wall-clock time. Given wide availability of distributed computing resources, all environment simulations at each iteration of training can be conducted totally in parallel. Thus, a more reasonable metric for the performance of a training algorithm is the number of non-parallelizable iterations, as opposed to the sample complexity. In this metric, ES is truly an appealing choice. 

Nevertheless, there are several challenges that need to be addressed in order to further improve the performance of ES in training complex policies. First, most ES methods cope with the non-smoothness of the objective function by considering a Gaussian-smoothed version of the expected total reward. The gradient of this function is intractable and must be estimated to provide the policy parameter update. In the pioneering work \citep{salimans2017evolution}, a gradient estimator is proposed based on random parameter sampling. Developing efficient sampling strategies for gradient estimates has become an interest in ES research since then, and several improvements have been proposed, based on imposing structures on parameter perturbation \citep{CRSTW18, Choromanski_ES-Active}, or reusing past evaluations, \citep{Choromanski_RBO,Maheswaranathan_GuidedES,Meier_OPTRL_2019}. Yet, most of these gradient estimators are of Monte Carlo type, therefore arguably affected by the low accuracy of Monte Carlo methods. For faster convergence of training (i.e., reducing the number of iterations), more accurate gradient estimators are desired, particularly in RL tasks where the policy has a large number of parameters to learn.
Another prominent challenge is that the landscape of the objective function is complex and possesses plentiful local maxima. There is a risk for any optimization algorithm to get trapped in some of those points and unable to explore the parameter space effectively. 
The Gaussian smoothing, with its ability to smooth out function and damps out small, insignificant fluctuations, is a strong candidate in this very challenge. Specifically, with a moderately large smoothing parameter (i.e., strong smoothing effect), we can expect the gradient of the smoothed objective function will be able to look outside unimportant variation in the adjacent area and detect the general trend of the function from a distance, therefore an efficient \textit{nonlocal} search direction. This potential of Gaussian smoothing, however, has not been explored in reinforcement learning.

In this paper, we propose a new strategy to accelerate the time-to-solution of reinforcement learning by exploiting the Directional Gaussian Smoothing Evolution Strategy (DGS-ES), recently developed in \citep{2020arXiv200203001Z}. The DGS-ES method introduced a new directional Gaussian smoothing (DGS) gradient operator, that smooths the original objective function only along $d$-orthogonal directions in the parameter space. In other words, the DGS gradient requires $d$ one-dimensional Gaussian convolutions, instead of one $d$-dimensional convolution in the existing ES methods. There are several advantages of using the DGS-gradient operator in reinforcement learning. First, each component of the DGS-gradient, represented as a one-dimensional integral, can be accurately approximated with various classic numerical integration techniques. When having Gaussian kernels, we use Gauss-Hermite quadrature rule which can provide spectral accuracy (see \cite{Handbook}) in the DGS gradient approximation. Second, the use of Gauss-Hermite quadrature also features embarrassing parallelism as the random sampling used in existing ES methods. Since the communication cost between computing processors/cores is neglectable, the total computing time for each iteration of training does not increase with the number of environment simulations given sufficient computing resources. Third, the directional smoothing approach enables nonlocal exploration which takes into account large variation of the objective function and disregards local fluctuations. This property will greatly help skipping local optima or saddle points during the training. It is demonstrated in \S \ref{sec:ex} that the proposed strategy can significantly reduce the number of iterations in training several benchmark reinforcement learning problems, compared to the state-of-the-art baselines. 

\subsection{RELATED WORKS}\label{sec:relate}
ES belongs to the family of blackbox optimization that employs random search techniques to optimize an objective function \citep{Rechenberg_ESbook_1973,Schwefel_1977}. The search can be guided via either covariance matrix adaptation \citep{Hansen_Ostermeier_CMA_01,Hansen_CMA}, or search gradients to be fed to first order optimization schemes, \citep{Wierstra_NES_14}. The DGS-ES method is in line with the second approach. 
Gaussian smoothed as a method for approximating search direction has been introduced in \citep{FKM05,NesterovSpokoiny15}. In recent years, ES has been revived as a popular strategy in machine learning. In addition to those mentioned above, applications of ES to RL can also be found in \cite{FGKM18,Ha_Schmidhuber,Houthooft18,muller2018challenges, liu2019trust,pourchot2018cem}. Applications of ES beyond RL include meta-learning \citep{Metz_NIPS18}, optimizing neural network architecture \citep{Real_ICML17,Miikkulainen_ICML17, Cui_NeurIPS18}, and direct training of neural network \citep{MorseStanley16}.   

Effective exploration is also critical for deep RL on high dimensional action and state spaces, and a wide variety of exploration strategies have been developed in recent years, to name a few, count-based exploration \citep{Ostrovski17, TangHouthooft17}, intrinsic motivation \citep{Bellemare16}, curiosity \citep{Pathak-ICML17} and variational information maximization \citep{Houthooft-VIME16}. For exploration techniques which add noise directly to the parameter space of policy, see \citep{Fortunato17, Plappert17}. Also, exploration can be combine with deep RL methods to improve sample efficiency \citep{Colas18}. Finally, for a recent survey on policy search, we refer to \citep{Sigaud-review19}.

\section{BACKGROUND}\label{sec:setting}
We introduce the background of reinforcement learning and discuss the application of ES methods in reinforcement learning.

\subsection{REINFORCEMENT LEARNING}\label{sec:RL}
Reinforcement learning considers agents that are able to take a sequence of actions in an environment, denoted by $\mathcal{E}$, over a number of discrete time steps $t \in \{1, \ldots, T\}$. At each time step $t$, the agent receives a state $\bm s_t \in \mathcal{S}$ and produces a follow-up action $\bm a_t \in \mathcal{A}$. The agent will then observe another state $\bm s_{t+1}$ and a scalar reward $r_t$. The goal of the agents is to learn a policy $\pi(\bm a_t | \bm s_t)$ that maximizes the objective function, i.e., the expected cumulative return of the form 
\begin{equation}\label{eq:return}
    J(\pi) := \sum_{t=0}^T \mathbb{E}_{(\bm s_t, \bm a_t)}\left[\gamma^t r(\bm s_t, \bm a_t)\right], 
\end{equation}
where $0 \le \gamma \le 1$ is a discount rate, $\bm a_t$ is drawn from the policy $\pi(\bm a_t | \bm s_t)$, and $\bm s_{t+1} = \mathcal{E}(\bm s_t, \bm a_t)$ is generated by running the environment dynamics. 

In policy-based reinforcement learning approaches, the policy $\pi(\bm a|\bm s; \bm \theta)$ is parameterized by $\bm \theta \in \mathbb{R}^d$, where the vector $\bm \theta:=(\theta_1, \ldots, \theta_d)^{\top}$ represents the parameters of the policy, e.g., weights of a neural network. Then, the task of learning a good policy $\pi$ becomes iterative updating the parameter $\bm \theta$ to solve the following optimization problem
\begin{equation}\label{eq:opt}
    \max_{\bm \theta \in \mathbb{R}^d} J(\bm \theta),
\end{equation}
where we denote $J(\bm \theta) := J(\pi(\bm a | \bm s; \bm \theta))$ by an abuse of notation. In reinforcement learning, it is usually true that the gradient of 
the environment $\mathcal{S}$ is inaccessible, so automatic differentiation cannot be used to obtain the gradient of $J(\bm \theta)$. Thus, much of the innovation in reinforcement learning algorithms is focused on addressing the lack of access to or the existence of gradients of the environment/policy. In this work, we focus on the evolution strategy and other types of training algorithms for reinforcement learning are summarized in \S\ref{sec:intro}.

\subsection{EVOLUTION STRATEGY}\label{sec:es}
We briefly recall the evolution strategy methods, e.g., \citep{Hansen_Ostermeier_CMA_01, salimans2017evolution}, which use a multivariate Gaussian distribution to generate the population around the current parameter value $\bm \theta_t$ at the $t$-iteration. When the Gaussian distribution can be factorized to $d$ independent one-dimensional Gaussian distributions, the standard ES method can be mathematically interpreted based on the Gaussian smoothing technique \citep{FKM05,NesterovSpokoiny15}. Specifically, a smoothed version of $J(\bm \theta)$ in Eq.~\eqref{eq:return}, denoted by $J_{\bm \sigma}(\bm \theta)$, is defined by 
\begin{align}
       J_{\bm \sigma}(\bm \theta) & := \frac{1}{(2\pi)^{\frac{d}{2}}} \int_{\mathbb{R}^d} J(\bm \theta + \bm \sigma \circ \bm u)\, {\rm e}^{-\frac{1}{2}\|\bm u\|^2_2}\, d\bm u \label{e30}\\
       &  = \mathbb{E}_{\bm u \sim \mathcal{N}(0, \mathbf{I}_d)} \left[J(\bm \theta + \bm \sigma \circ \bm u) \right] \notag,
\end{align}
where $\mathcal{N}(0, \mathbf{I}_d)$ is a $d$-dimensional standard Gaussian distribution, the notation ``$\circ$'' represents the element-wise
product, and the vector $\bm \sigma = (\sigma_1, \ldots, \sigma_d)$ controls the smoothing effect. It is well known that $J_{\bm \sigma}$ is always differentiable even if $J$ is not. In addition, most of the characteristics of the original objective function $J(\bm \theta)$, e.g., convexity, the Lipschitz constant, are inherited by $J_{\bm \sigma}(\bm \theta)$. 
%
%
When $J(\bm \theta)$ is differentiable, the difference $\nabla J - \nabla J_{\bm \theta}$ can be bounded by its Lipschitz constant (see \citep{NesterovSpokoiny15}, Lemma 3 for details). Thus, the original optimization problem in Eq.~(\ref{eq:opt}) can be replaced by a smoothed version, i.e.,
\begin{equation}\label{eq:smooth}
    \max_{\bm \theta \in \mathbb{R}^d} J_{\bm \sigma}(\bm \theta),
\end{equation}
where the gradient of $J_{\bm \sigma}(\bm \theta)$ is given by
\begin{equation}\label{e40}
    \nabla J_{\bm \sigma}(\bm \theta)  
    = \frac{1}{\|\bm \sigma\|_2^{d/2}}\mathbb{E}_{\bm u \sim \mathcal{N}(0, \mathbf{I}_d)} \left[J(\bm \theta + \bm \sigma \circ \bm u)\, \bm u\right].
\end{equation}
The standard ES method \citep{salimans2017evolution} uses Monte Carlo sampling to estimate the gradient $\nabla J_{\bm \sigma}(\bm \theta)$ and update the state $\bm \theta$ from iteration $n$ to $n+1$ by 
\begin{equation}\label{eq:ES}
\bm \theta_{n+1} = \bm \theta_n - \frac{\lambda}{M\sigma}\sum_{m=1}^M J(\bm \theta_n + \bm \sigma \circ \bm u_m) \bm u_m,
\end{equation}
where $\lambda$ is the learning rate, $\bm u_m$ are sampled from the Gaussian distribution $\mathcal{N}(0,\mathbf{I}_d)$. 

One drawback of the ES method and its variants is the slow convergence of the training process, due to the low accuracy of the MC-based gradient estimator {(see \citep{BCCS19}), also \citep{2020arXiv200203001Z}, for extended discussions on the accuracy of gradient approximations using Eq.~\eqref{eq:ES} and related methods).} On the other hand, the evaluations of $J(\bm \theta_n + \bm \sigma \circ \bm u_m)$ for $m = 1, \ldots, M$ at the $n$-th iteration can be generated totally in parallel, which makes it well suited to be scaled up to a large number of parallel workers on modern supercomputers. Therefore, \emph{the motivation of this work is to develop a new gradient operator to replace the one in Eq.~\eqref{e40}, such that the new gradient can be approximated in a much more accurate way and the embarrassing parallelism feature can be retained.}

\section{THE DGS-ES METHOD}
\label{sec:DGS-ES}
This section introduces our main framework. 
We start by introducing in \S \ref{sec:grad} the DGS gradient operator and its approximation using the Gauss-Hermite quadrature rule. In \S \ref{sec:ada_DGS-ES}, we describe how to incorporate the DGS gradient operator into the ES for reinforcement learning.

\subsection{THE DGS GRADIENT OPEARTOR}\label{sec:grad}
For a given direction $\bm \xi \in \mathbb{R}^d$, the restriction of the objective function $J(\bm \theta)$ along $\bm \xi$ can be represented by
\begin{equation}\label{e9}
G(y \,| \,{\bm \theta, \bm \xi}) = J(\bm \theta + y\, \bm \xi), \;\; y \in \mathbb{R},
\end{equation}
where $\bm \theta$ is the current state of the agent's parameters. Then, 
we can define the one-dimensional Gaussian smoothing of $G(y)$, denoted by $G_\sigma(y)$, by
\begin{equation}
\label{eq10}
\begin{aligned}
     G_{\sigma}(y \,| \,{\bm \theta, \bm \xi})  := \mathbb{E}_{v \sim \mathcal{N}(0, 1)} \left[G(y + \sigma v\, |\, \bm \theta, \bm \xi) \right],
\end{aligned}
\end{equation}
which is also the Gaussian smoothing of $J(\bm \theta)$ along $\bm \xi$ in the neighbourhood of $\bm \theta$. The derivative of $G_{\sigma}(y|\bm \theta,\bm \xi)$ at $y = 0$ is given by
\begin{equation}\label{e4}
\begin{aligned}
    \mathscr{D}[G_{\sigma}(0 \,|\, \bm \theta, \bm \xi)]
    = \frac{1}{\sigma}\,\mathbb{E}_{v \sim \mathcal{N}(0,1)} \left[G(\sigma v \, | \, \bm \theta, \bm \xi)\, v\right],
\end{aligned}
\end{equation}
where $\mathscr{D}$ denotes the differential operator. 
It is easy to see that 
 $\mathscr{D}[G_{\sigma}(0 \,|\, \bm x, \bm \xi)]$ 
only involves the directionally smoothed objective function given in Eq.~\eqref{eq10}. 

We can assemble a new gradient, i.e., the DGS gradient, by putting together the  derivatives in Eq.~\eqref{e4} along $d$ orthogonal directions, i.e., 
\begin{equation}\label{dev_smooth_func}
    {\nabla}_{\bm \sigma, \bm \Xi}[J](\bm \theta) = \bm \Xi^{\top}
\begin{bmatrix}
{\mathscr{D}}\left[G_{\sigma_1}(0 \, |\, \bm \theta, \bm \xi_1)\right]\\
\vdots \\
{\mathscr{D}}\left[G_{\sigma_d}(0\, |\, \bm \theta, \bm \xi_d) \right]
\end{bmatrix},
\end{equation}
where $\bm \Xi := (\bm \xi_1, \ldots, \bm \xi_d)^{\top}$ represents the matrix consisting of 
$d$ orthonormal vectors. It is important to notice that
\[
\nabla J_{\bm \sigma}(\bm \theta) \not= {\nabla}_{\bm \sigma, \bm \Xi}[J](\bm \theta)
\]
for any $\bm \sigma >0$, because of the directional Gaussian smoothing used in Eq.~\eqref{dev_smooth_func}.
%
However, there is consistency between the two quantities as $\bm\sigma \rightarrow 0$, i.e.,
\begin{equation}\label{consist}
\lim_{\bm \sigma \rightarrow 0} \left| \nabla J_{\bm \sigma}(\bm \theta) - {\nabla}_{\bm \sigma, \bm \Xi}[J](\bm \theta) \right| = 0,
\end{equation}
for fixed $\bm \theta$ and $\bm \Xi$. If $\nabla J(\bm \theta)$ exists, then ${\nabla}_{\bm \sigma, \bm \Xi}[J](\bm \theta)$ will also converge to $\nabla J(\bm \theta)$ as $\bm \sigma \rightarrow 0$.
%
Such consistency naturally led to the idea of replacing $\nabla J_{\bm \sigma}(\bm \theta)$ with ${\nabla}_{\bm \sigma, \bm \Xi}[J](\bm \theta)$ in the ES framework. 


\subsection{THE DGS-ES ALGORITHM FOR REINFORCEMENT LEARNING}\label{sec:ada_DGS-ES}
Since each component of 
${\nabla}_{\bm \sigma, \bm \Xi}[J](\bm \theta)$ in Eq.~\eqref{dev_smooth_func} only involves a one-dimensional integral, we can use Gaussian quadrature rules  \citep{2013JSV...332.4403B} to obtain spectral convergence. In the case of Gaussian smoothing, a natural choice is the Gauss-Hermite (GH) rule, 
%
which is used to approximate integrals of the form $\int_{\mathbb{R}} g(x) {\rm e}^{-x^2}dx$. By doing a simple change of variable in Eq.~\eqref{e4}, the GH rule can be directly used to obtain the following estimator:
\begin{equation}\label{e8}
\begin{aligned}
    & \widetilde{\mathscr{D}}^M[G_\sigma(0 \, | \, \bm \theta, \bm \xi)] \\ := &\frac{1}{\sqrt{\pi}\sigma} \sum_{m = 1}^M w_m \,G(\sqrt{2}\sigma v_m \,|\, \bm \theta, \bm \xi)\sqrt{2}v_m\\
\end{aligned}   
\end{equation}
where $w_m$ are the GH quadrature weights defined by 
\begin{equation}\label{eq:GHw}
w_m=\frac{2^{M+1}M\,!\sqrt{\pi}}{[H_M'(v_m)]^2},\;\; m = 1, \ldots, M,
\end{equation}
$v_m$ are the roots of the Hermite polynomial of
degree $M$
\begin{equation}\label{eq:GHx}
   H_M(v)=(-1)^M {\rm e}^{v^2}\frac{d^M}{dv^M}({\rm e}^{-v^2}),
\end{equation}
 and $M$ is the number of function evaluations, i.e., environment simulations, needed to compute the quadrature in Eq.~\eqref{e8}. The weights $\{w_m\}_{m=1}^M$ and the roots $\{v_m\}_{m=1}^M$
can be found 
in \cite{Handbook}.
The approximation error of the GH formula can be bounded by $\widetilde{\mathscr{D}}^M[G_\sigma]$ is
\begin{align}
\label{GH_error}
\left|\widetilde{\mathscr{D}}^M[G_\sigma] - \mathscr{D}[G_\sigma] \right| \le C_0\frac{M\,!\sqrt{\pi}}{2^M(2M)\,!} \sigma^{2M-1}, 
\end{align}
where $M!$ is the factorial of $M$, and the constant $C_0>0$ is independent of $M$ and $\sigma$. 

Applying the GH quadrature rule $\widetilde{\mathscr{D}}^M$ to each component of ${\nabla}_{\bm \sigma, \bm \Xi}[J](\bm \theta)$ in Eq.~\eqref{dev_smooth_func}, we define the following estimator: 
\begin{equation}\label{e5}
    \widetilde{\nabla}^M_{\bm \sigma, \bm \Xi}[J](\bm \theta) := \bm \Xi^{\top}
\begin{bmatrix}
\widetilde{\mathscr{D}}^M\left[G_{\sigma_1}(0 \, |\, \bm \theta, \bm \xi_1)\right]\\
\vdots \\
\widetilde{\mathscr{D}}^M\left[G_{\sigma_d}(0\, |\, \bm \theta, \bm \xi_d) \right]
\end{bmatrix},
\end{equation}
which requires a total of $M\times d$ \emph{parallelizable} environment simulations at each iteration of training. %
The error bound in Eq.~\eqref{GH_error} indicates that $\widetilde{\nabla}^M_{\bm \sigma, \bm \Xi}[J](\bm \theta)$ is an accurate estimator of the DGS gradient for a small $M$, regardless of the value of $\bm \sigma$. This enables the use of relatively big values of $\bm \sigma$ in Eq.~\eqref{e5} to exploit the \emph{nonlocal} features of the landscape of $J(\bm \theta)$ in the training process. The nonlocal exploitation ability of $\widetilde{\nabla}^M_{\bm \sigma, \bm \Xi}[J]$ is demonstrated in \S\ref{sec:ex} to be effective in reducing the necessary number of iterations to achieve a prescribed reward score.

On the other hand, as the quadrature weights $w_m$ and $v_m$ defined in Eq.~\eqref{eq:GHw} and Eq.~\eqref{eq:GHx}, are deterministic values, the DGS estimator $\widetilde{\nabla}^M_{\bm \sigma, \bm \Xi}[J](\bm x)$ in Eq.~\eqref{e5} is also a \emph{deterministic} for fixed $\bm \Xi$ and $\bm \sigma$. To introduce random exploration ability to our approach, we add random perturbations to both $\bm \Xi$ and $\bm \sigma$. 
For the orthonormal matrix $\bm \Xi$, we add a small random rotation, 
denoted by $\Delta \bm\Xi$, to the current matrix $\bm \Xi$. The matrix $\Delta \bm\Xi$ is generated as a random skew-symmetric matrix, of which the magnitude of each entry is smaller than a prescribed threshold $\alpha >0$. The perturbation of $\bm \sigma$ is conducted by drawing random samples from a uniform distribution $U(r-\beta, r+\beta)$ with two hyper-parameters $r$ and $\beta$ with $r-\beta>0$. The random perturbation of $\bm \Xi$ and $\bm \sigma$ can be triggered by various types of indicators e.g., the magnitude of the DGS-gradient, the number of iteration done since last perturbation.

\vspace{-0.0cm}
\begin{algorithm}[h!]
\setstretch{1.1}
  \caption{\hspace{-0.1cm}: The DGS-ES for reinforcement learning}
  \label{alg1}
\begin{algorithmic}[1]
  \STATE{\bf Hyper-parameters}: \\
  $M$: the order of GH quadrature rule\\
  $\alpha$: the scaling factor for controlling the norm of $\Delta \bm \Xi$\\
  $r, \beta$: the mean and perturbation scale for sampling $\bm \sigma$\\
  $\gamma$: the tolerance for triggering random perturbation
  \STATE{\bfseries Input:}\\ $\bm \theta_0$: the initial parameter value,\\ 
  $L$: the number of parallel workers
  \STATE{\bfseries Output:} the final parameter value $\bm \theta_N$
  \STATE Initialize the policy $\pi$ with $\bm \theta_0$
  \STATE Set $\bm \Xi = \mathbf{I}_d$, and $\sigma_i = r$ for $i = 1, \ldots, d$
  \STATE Broadcast $L$ copies of $\pi(\bm a|\bm s; \bm \theta_0)$ to the $L$ workers
  \STATE Divide the total GH quadrature points into $L$ subsets, and send each subset to a worker.
  \FOR{$n=0, \ldots N-1$}
  \STATE Each worker runs $Md/L$ environment simulations at their assigned quadrature points
  \STATE Each worker sends $Md/L$ scores to the master 
  %
  \FOR{$i = 1, \ldots, d$}
  \STATE Compute $\widetilde{\mathscr{D}}^M[G_{\sigma_i}(0\,|\, \bm \theta_n, \bm \xi_i)]$ in Eq.~(\ref{e8})
  \ENDFOR
  \STATE Assemble $\widetilde{\nabla}^M_{\bm \sigma, \bm \Xi}[J](\bm \theta_n)$ in Eq.~(\ref{e5})
  \STATE Update $\bm \theta_n$ to $\bm \theta_{n+1}$ using Adam
\vspace{0.05cm}
  \IF{$\|\widetilde{\nabla}^M_{\bm \sigma, \bm \Xi}[J](\bm \theta_n)\|_2 < \gamma$}
  \vspace{0.05cm}
  \STATE Generate $\Delta \bm \Xi$ and update $\bm \Xi = \mathbf{I}_d + \Delta \bm \Xi$
  \STATE Generate $\bm \sigma$ from $U(r-\beta, r+\beta)$
  \ENDIF
  \STATE Broadcast $\bm \theta_{n+1}$ to the $L$ workers
  \STATE Each worker updates the policy to $\pi(\bm a|\bm s; \bm \theta_{n+1})$
  \ENDFOR
\end{algorithmic}
\end{algorithm}

\section{EXPERIMENTS}\label{sec:ex}

To evaluate the DGS-ES algorithm, we tested its performance on two classes of reinforcement learning environments: three classical control theory problems from OpenAI Gym (\url{https://github.com/openai/gym}) \citep{brockman2016openai} and three continuous control tasks simulated using PyBullet (2.6.5)  (\url{https://pybullet.org/}) \citep{coumans2016pybullet} which is an open-source library. 
Within OpenAI Gym, we demonstrate the proposed approach on three benchmark examples: {\color{mydarkblue}CartPole-v0} (discrete), {\color{mydarkblue} MountainCarContinuous-v0} (continuous), {\color{mydarkblue}Pendulum-v0} (continuous). The maximum time steps for these examples are 200, 999 and 200, respectively. More details about the environment and reward settings can be found in \cite{brockman2016openai}. We also examined the DGS-ES algorithm on the challenging continuous robotic control problems in PyBullet library, namely {\color{mydarkblue}HopperBulletEnv-v0}, {\color{mydarkblue}InvertedPendulumBulletEnv-v0} and {\color{mydarkblue}ReacherBulletEnv-v0}. In these three tasks, the maximum time steps are 1000, 1000 and 150, respectively. For the purpose of reproducible comparison, we employed the original environment settings from the OpenAI Gym and the PyBullet library without modifying the rewards or the environments. 

For our implementation of DGS-ES, we defined our policies as a two-layer feed-forward neural network with 16 hidden nodes and tanh activation functions. For gradient-based optimization, we used Adam to adaptively update the network parameters with a learning rate of $\ell_r=0.1$. We chose the hyper-parameters used in Algorithm 1 as follows: $M=7, \alpha=2.0, r=1.0, \beta=0.2, \gamma=0.01$. In practice one can tune the critical hyper-parameters ($M, r$ and $\ell_r$) given the following suggested range: $M \in [7,9]$, $r \in [0.5,1.0]$ and $\ell_r \in [0.01, 0.1]$. For each task, our results were performed over 5 repeated independent trials (different random seeds) of the Gym/PyBullet simulators and the network policy initialization. 

The DGS-ES algorithm is specifically amenable to parallelization since it only needs to communicate scalars, allowing it to scale to over a large number of parallel workers. We implemented a distributed version of Algorithm 1 to the reinforcement learning tasks. The distributed DGS-ES is implemented using PyTorch \citep{paszke2017automatic} combined with Ray \citep{moritz2018ray} (\url{https://github.com/ray-project/ray}), which does not rely on special networking setup and was tested on large-scale high performance computing facilities with thousands of computing nodes/workers. 

{\bf Comparison metric}: As the motivation of this work is to accelerate time-to-solution of reinforcement training under the assumption that sufficient distributed computing resource is available, we used a different metric to evaluate the performance of DGS-ES and the baselines. Specifically, we are interested in the average return $\mathbb{E}[J]$ versus the number of iterations, i.e., $N$ in Algorithm 1, because those iterations cannot be parallelized.

\subsection{BASELINE METHODS}
We compared Algorithm 1 against several RL baselines, including ES, PPO and TRPO, as well as the state-of-the-art algorithms such as ASEBO, DDPG, and TD3. Below is the information of the packages used in this effort.
\vspace{-0.2cm}
\begin{itemize}
    \item {\bf ES}: The Evolution Strategy proposed in \cite{salimans2017evolution}. We used the implementation of ES from the open-source code \url{https://github.com/hardmaru/estool}.
    \item {\bf ASEBO}: Adaptive ES-Active Subspaces for Blackbox Optimization, which was recently developed by \cite{Choromanski_ES-Active}. We used the implementation released by the authors at \url{https://github.com/jparkerholder/ASEBO}. 
    \item {\bf PPO}: Proximal Policy Optimization in \cite{Schulman-PPO}, which is available in OpenAI's baselines repository at \url{https://github.com/openai/baselines} \citep{baselines}.
    \item {\bf TRPO}: Trust Region Policy Optimization, developed by \cite{Schulman_TRPO}. We also used the OpenAI's baselines implementation \citep{baselines}.  
    \item {\bf DDPG}: Deep Deterministic Policy Gradient, proposed by \cite{Lillicrap-DDPG}. We used the implementation from \url{https://github.com/georgesung/TD3} where the benchmark DDPG in PyBullet is provided. 
    \item {\bf TD3}: Twin Delayed Deep Deterministic Policy Gradient \citep{fujimoto2018addressing}, which was built upon the DDPG. The original results were reported for the MuJoCo version environments using the implementation from \url{https://github.com/sfujim/TD3}, but we used the PyBullet implementation from \url{https://github.com/georgesung/TD3}. 
\end{itemize}

The hyper-parameters for all algorithms above were set to match the original papers without further tuning to improve performance on the testing benchmark examples.

\subsection{COMPARATIVE EVALUATION}
\begin{figure}[h!]
     \centering
  \includegraphics[scale = 0.191]{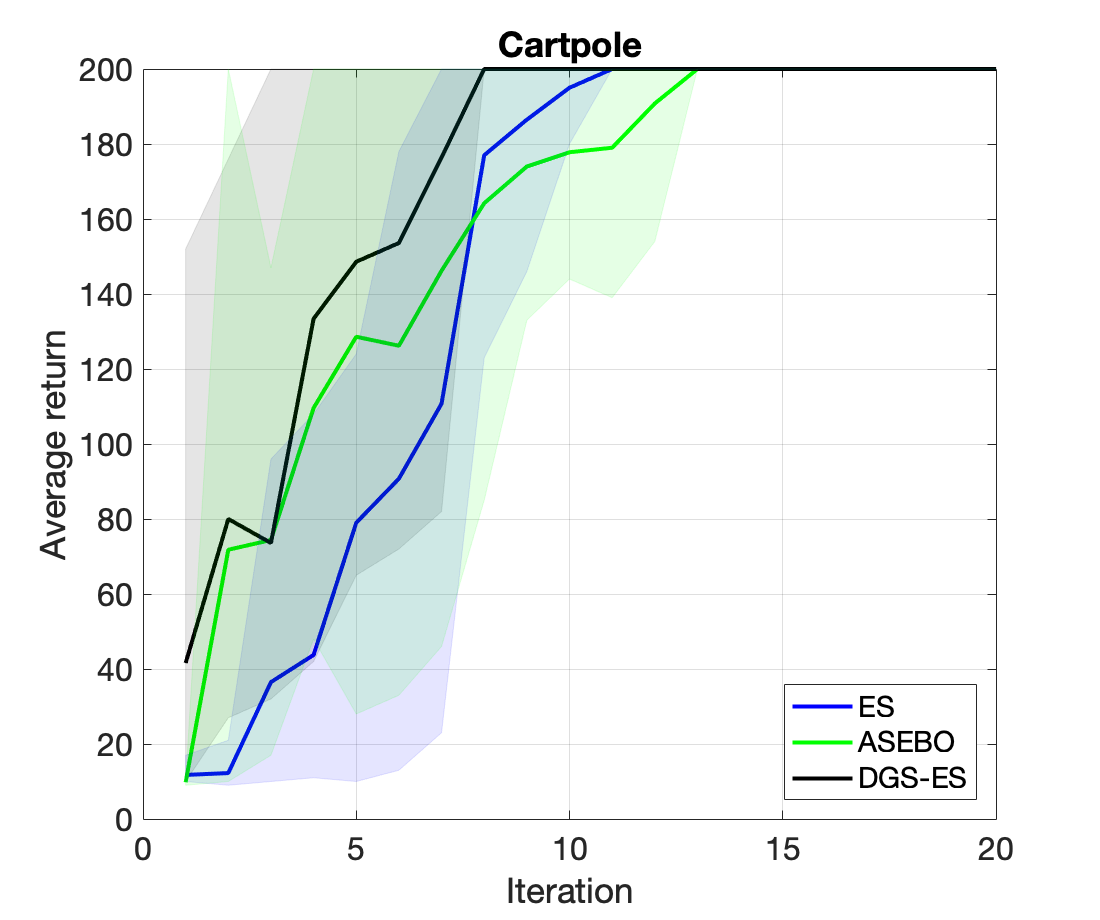}
    \includegraphics[scale = 0.191]{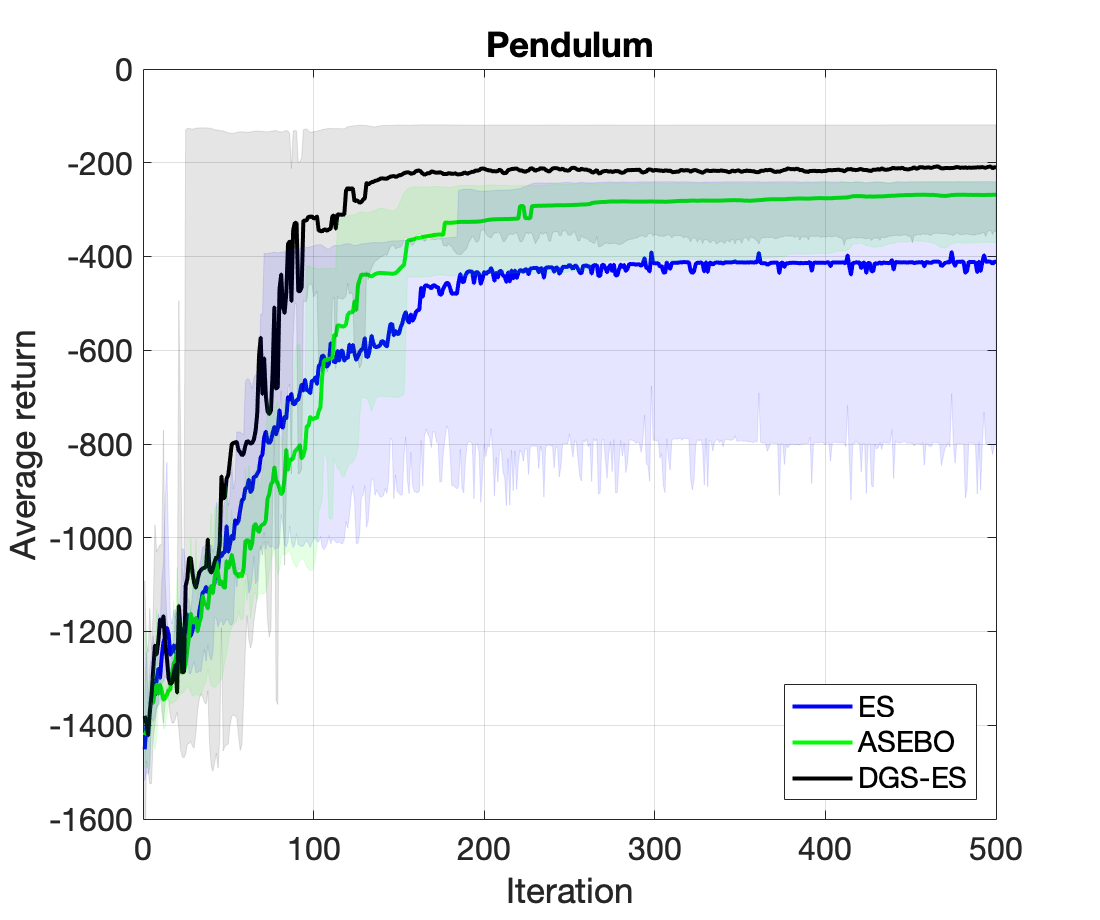}
    \includegraphics[scale = 0.191]{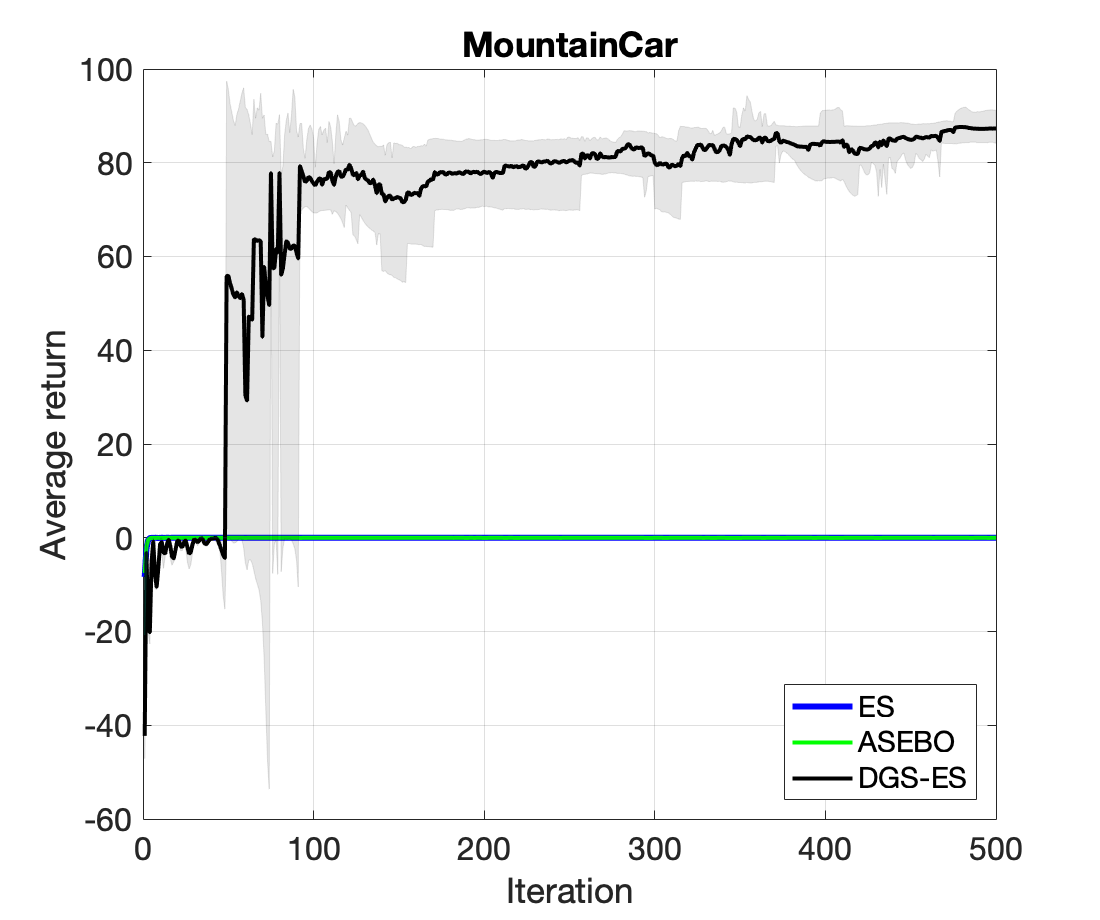}
    \caption{Comparison between the DGS-ES and two baselines, i.e., ES and ASEBO, for solving the three problems from OpenAI Gym. The colored curves are the average return over 5 repeated runs with different random seeds, and the corresponding shade represents the interval between the maximum and minimum return.}
    \label{Pendulum}
\end{figure}

Figure \ref{Pendulum} shows the comparison results of CartPole, Pendulum and MountainCar problems from the OpenAI Gym. We compared the DGS-ES with classical ES and the improved ASEBO method. In general, the DGS-ES method features faster convergence than the baselines. For the simplest CartPole problem, the three methods perform equally well. Discrepancy appears in the Pendulum test, where the DGS-ES method not only converges faster than the baselines, but also achieves a higher average return. There is a much bigger discrepancy between the DGS-ES and the baselines appear in the MountainCar test. According to the guideline provided in the OpenAI Gym, the success threshold is to achieve an average return of 90. The DGS-ES method achieves the threshold within 500 iterations, while the average returns of the ES and ASEBO methods are around zero. It is well known that the challenge of this problem is that the surface of the objective function $J(\bm \theta)$ is very flat at most locations in the parameter space, which makes it difficult to capture the peak of $J(\bm \theta)$. This test is a good demonstration of the nonlocal exploration ability of the DGS-ES method. 
Since the mean of the smoothing factor $\bm \sigma$ is set to 1.0, 
DGS-ES can capture the peak of $J(\bm \theta)$ much faster than the baselines. In fact, we can see that it took around 50 iterations for the DGS-ES to find the peak region, while ES and ASEBO needed more than 3000 iterations (not plotted) to move out of the flat region. 

\begin{figure}[h!]
     \centering
  \includegraphics[scale = 0.19]{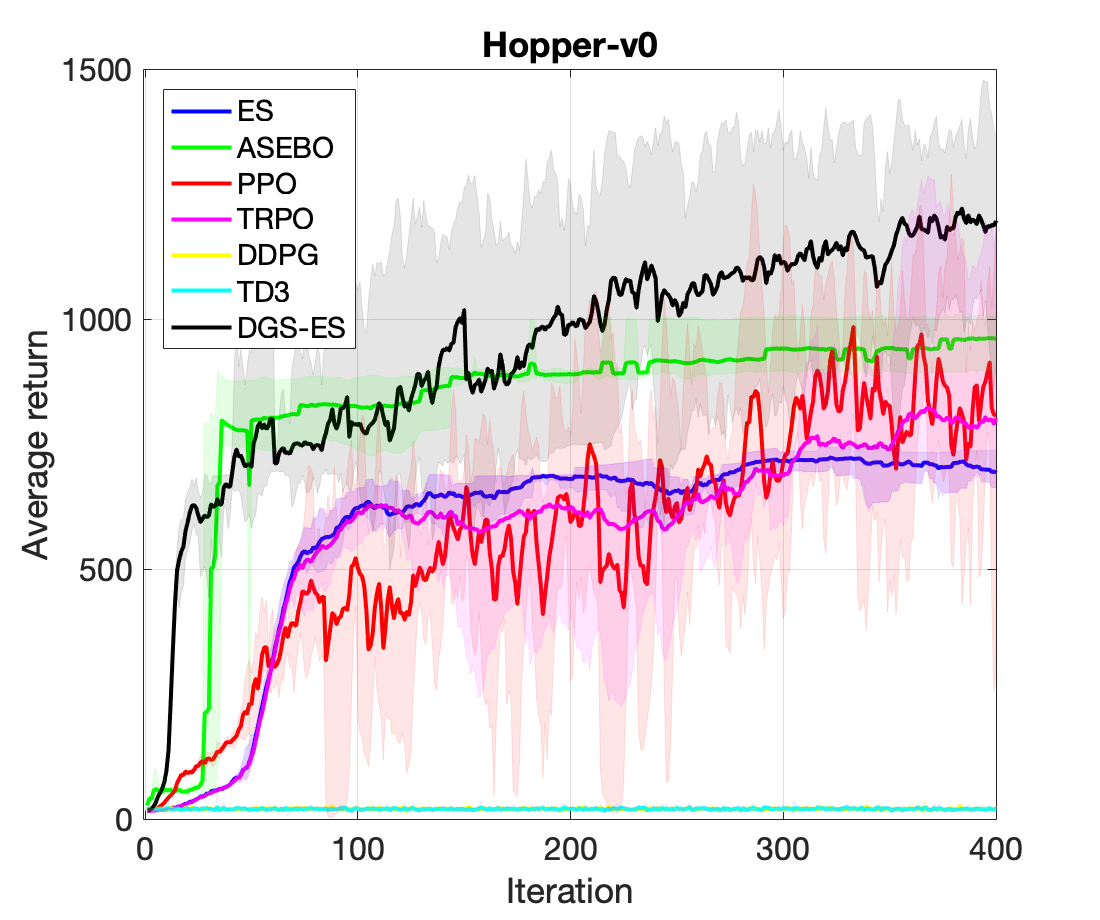}
   \includegraphics[scale = 0.19]{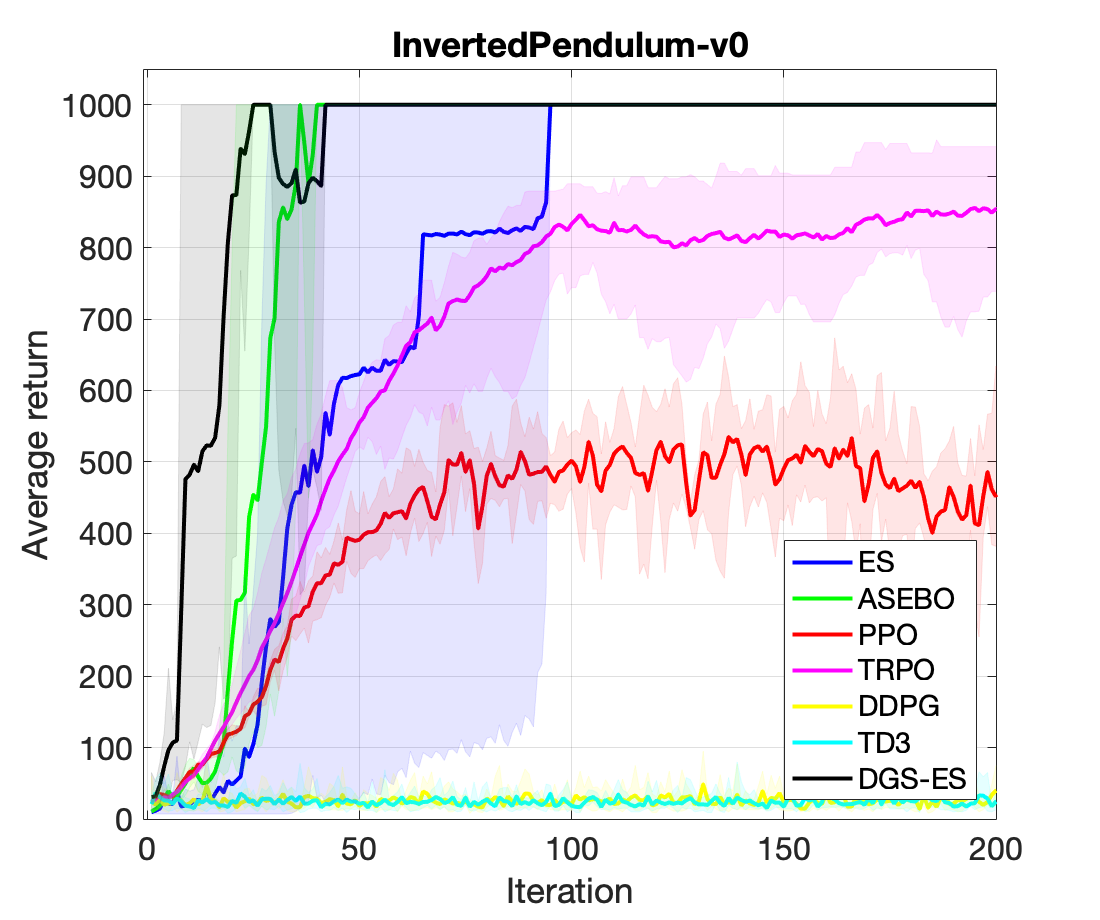}
    \includegraphics[scale = 0.19]{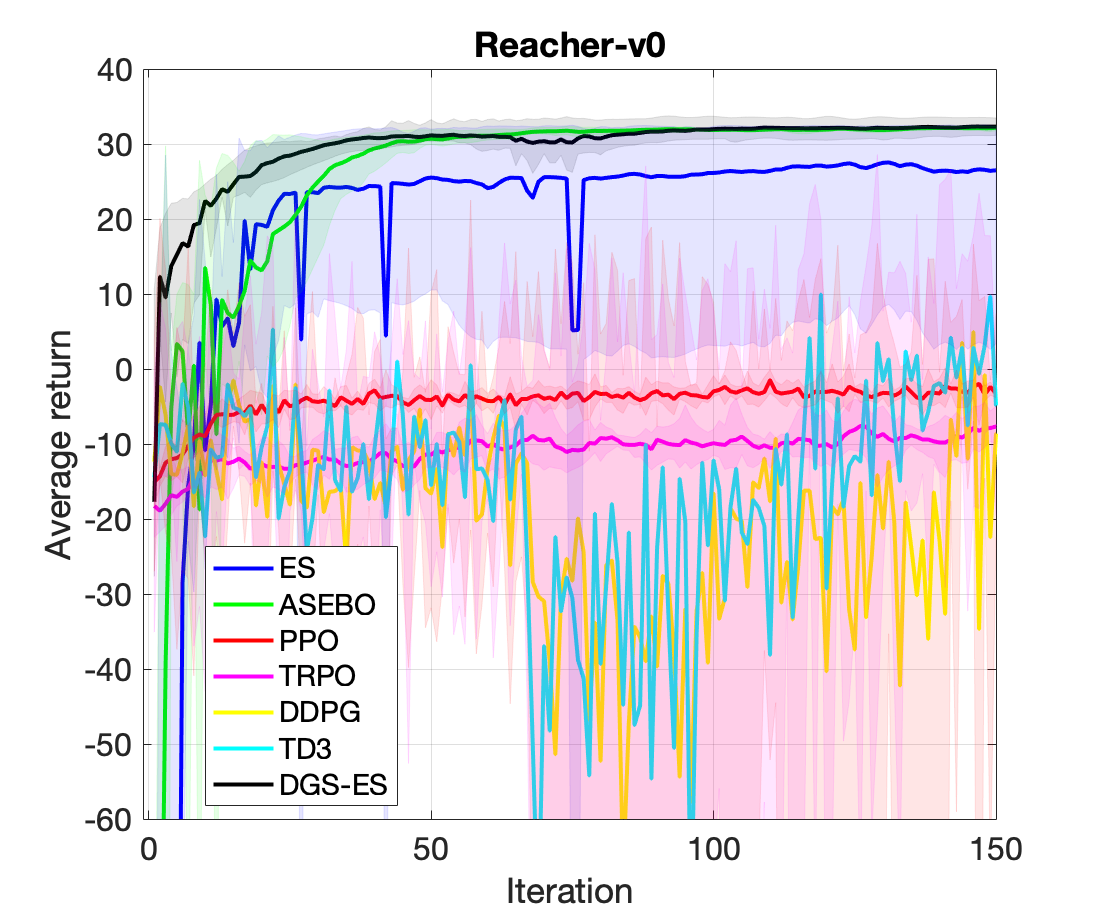}
    \caption{Comparison between the DGS-ES method and the baselines, i.e., ES, ASEBO, PPO, TRPO, DDPG and TD3, for solving the three problems from the PyBullet library. The colored curves are the average return over 5 runs with different random seeds, and the corresponding shade represents the interval between the maximum and minimum return.} 
    \label{Hopper}
\end{figure}

Figure \ref{Hopper} shows the comparison results of Hopper-v0, InvertedPendulum-v0 and Reacher-v0 problems from the
 PyBullet library. We compared the DGS-ES with six baselines including ES, ASEBO, PPO, TRPO, DDPG and TD3. As expected, the DGS-ES method shows better performance in terms of the convergence speed. For the Hopper-v0 problem, the DGS-ES achieves the highest return of all the methods within 400 iterations. It is worth pointing out that some of the baselines will eventually catch up with the DGS-ES given a sufficiently large number of iterations. For example, DDPG and TD3 do not provide much improvement within 400 iterations, but DDPG could reach 1650 average return with over 3000 
iterations, and TD3 could reach even higher, around 2200 average return, with over 6000 iterations, according to the baselines provided by OpenAI \citep{baselines} and \cite{fujimoto2018addressing}. This phenomenon illustrates the fast convergence feature of DGS-ES. 
 For the InvertedPendulum-v0 problem,  DGS-ES can achieve the maximum return 1000 (default value in the PyBellet library) around 30 iterations. In comparison, ES and ASEBO can reach the maximum return but with more iterations than DGS-ES. According to the benchmark results for PyBullet environments in \cite{fujimoto2018addressing}, DDPG and TD3 cannot converge to the maximum return even with a large number of iterations. For the Reacher-v0 problem, DGS-ES and ASEBO are still the top performers, and the advantage of DGS-ES is, again, faster convergence.

\begin{figure}[h!]
     \centering
  \includegraphics[scale = 0.2]{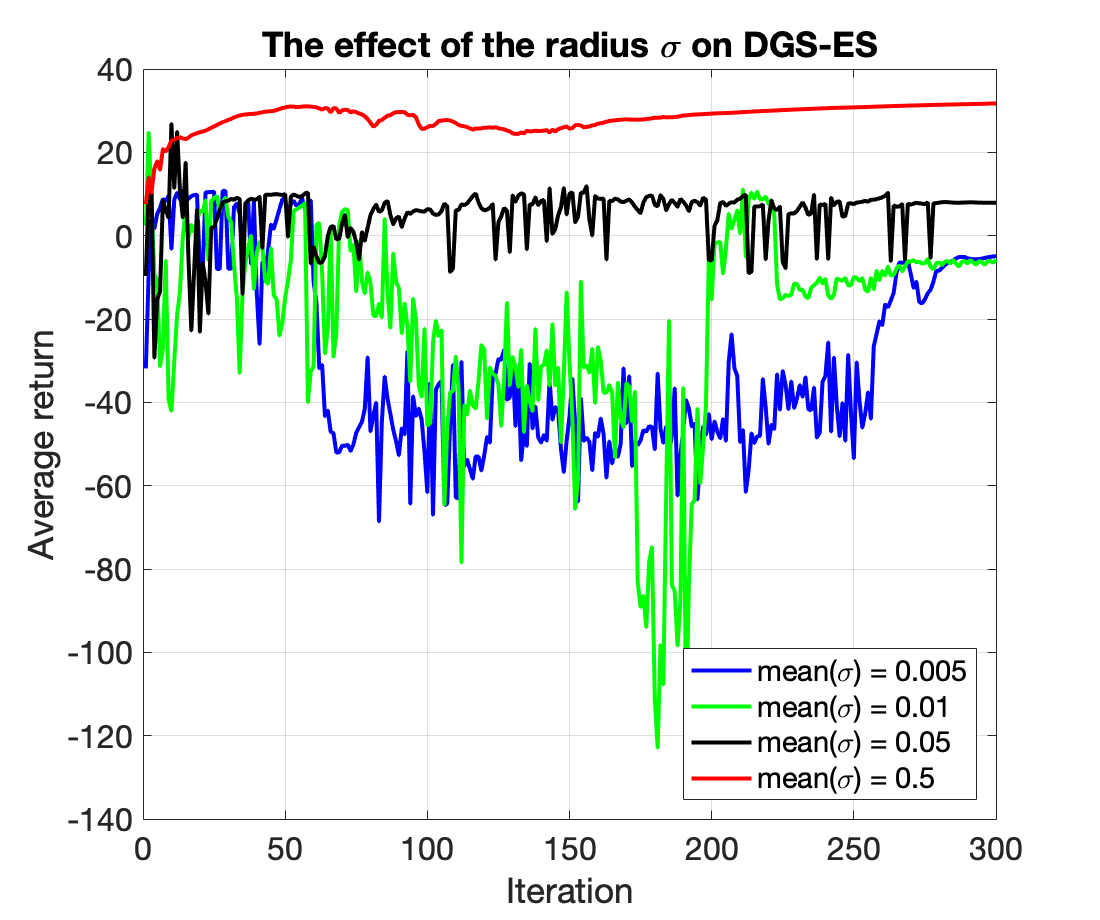}
  \vspace{-0.2cm}
    \caption{Illustration of the effect of the radius $\bm \sigma$ of $\widetilde{\nabla}^M_{\bm \sigma, \bm \Xi}[J](\bm \theta)$ on the performance of the DGS-ES method in solving the Reacher-v0 problem.}
    \label{sigma}
\end{figure}

Figure \ref{sigma} illustrates the effect of the radius $\bm \sigma$ of $\widetilde{\nabla}^M_{\bm \sigma, \bm \Xi}[J](\bm \theta)$ on the performance of the DGS-ES method in solving the Reacher-v0 problem. All the simulations were done using the same initialization. We set the mean of , i.e., the hyper-parameter $r$ in Algorithm 1, to 0.5, 0.05, 0.01, and 0.005. It is easy to see that the performance of DGS-ES deteriorates with the decrease of 
$\bm \sigma$. Since it is evident that the surface of $J(\bm \theta)$ is not convex and may have many local maxima, a relatively big radius $\bm \sigma$ is necessary to help DGS-ES skip the local maxima. As $\bm \sigma$ becomes smaller, the DGS gradient $\widetilde{\nabla}^M_{\bm \sigma, \bm \Xi}[J](\bm \theta)$ converges to the local gradient $\nabla J(\bm \theta)$, which may get the optimizer trapped in a local maximum. 

\section{CONCLUSION}\label{sec:con}
Despite the successful demonstration shown in \S \ref{sec:ex}, there are several limitations with the DGS-ES method for reinforcement learning. First, it requires a powerful enough cluster or distributed computing resources to show superior performance. Even though more and more parallel environments for complex RL tasks, those parallel codes might not compatible with a cluster/supercomputer with a specific architecture. This will need some extra effort to modify environment codes, in order to exploit the advantage of the DGS-ES approach. Second, asynchronization between different environment simulations may drag down the total performance. In a distributed computing system, all the parallel workers receive the same number of environment simulations. However, as different parameter values may lead to different termination times of the environment simulations, there will be a potential waste of computing resources due to such asynchronization. Thus, a better scheduling algorithm is needed to further improve the performance of DGS-ES in RL. Third, even though the performance of the DGS-ES is not very sensitive to the hyper-parameters, especially the radius $\bm \sigma$ in the experiments conducted in this work, optimal or even viable hyper-parameters of the DGS-ES method are still problem dependent, which means hyper-parameter tuning may be needed when applying the method to another RL problem.

\section*{Acknowledgement}
This material was based upon work supported by the U.S. Department of Energy, Office of Science, Office of Advanced Scientific Computing Research, Applied Mathematics program under contract and award numbers ERKJ352, by the DOE SciDac FastMath project, and by the Artificial Intelligence Initiative at the Oak Ridge National Laboratory (ORNL). ORNL is operated by UT-Battelle, LLC., for the U.S. Department of Energy under Contract DE-AC05-00OR22725.

%

\end{document}